\documentclass[11pt,a4paper]{article}
\usepackage[hyperref]{acl2020}
\usepackage{times}
\usepackage{booktabs}
\usepackage{array}
\usepackage{multirow}

\usepackage{graphicx}
\usepackage[linesnumbered,ruled,vlined]{algorithm2e}
\usepackage{url}
\usepackage{framed}
\usepackage{wrapfig}
\usepackage{tablefootnote}
\usepackage{amsmath}
\aclfinalcopy 
\def\aclpaperid{2878} 

\title{Exploring Content Selection in Summarization of Novel Chapters} 

\author{Faisal Ladhak\textsuperscript{1}\Thanks{  Equal contribution. Work done while at Amazon.} , Bryan Li\textsuperscript{2}\footnotemark[1] , Yaser Al-Onaizan\textsuperscript{3}, Kathleen McKeown\textsuperscript{1,3} \\
  \textsuperscript{1}Columbia University, \textsuperscript{2}University of Pennsylvania, \textsuperscript{3}Amazon AI\\
  \texttt{faisal.ladhak@columbia.edu, bryanli@seas.upenn.edu, }\\
  \texttt{onaizan@amazon.com, kathy@cs.columbia.edu}
  }
\date{}

\begin{document}
\maketitle
\begin{abstract}
We present 
a new summarization task, 
generating summaries of novel chapters  
using summary/chapter pairs from online study guides. 
This is a
harder task than the  news summarization task, given the chapter length as well as the extreme paraphrasing and generalization found in the summaries. 
We focus on extractive summarization, which
requires the creation of a gold-standard set of extractive summaries. We present a new metric for aligning  reference summary sentences with chapter sentences to create gold extracts and also experiment with different alignment methods. Our experiments demonstrate significant improvement over prior alignment approaches for our task  as shown through automatic metrics and a crowd-sourced pyramid analysis.  We make our data collection scripts available at \url{https://github.com/manestay/novel-chapter-dataset}.
\end{abstract}

\section{Introduction}

When picking up a novel one is reading, it would be helpful to be reminded of what happened last. To address this need, we develop an approach to generate extractive summaries of novel chapters. This is much harder than the news summarization tasks on which most of the summarization field (e.g., \cite{cheng-lapata-2016-neural,grusky-etal-2018-newsroom,paulus2017deep}) focuses; chapters are on average seven times longer than news articles.
There is no one-to-one correspondence between summary and chapter sentences, and the summaries in our dataset use extensive paraphrasing, while news summaries copy most of their information from the words used in the article.

We focus on the task of content selection, taking an initial, extractive summarization approach
given the task difficulty.\footnote{We tried two abstractive models~\cite{chen2018fast,liu-lapata-2019-text} but ROUGE was low and the output was poor with many repetitions and hallucinations.} 
As the reference summaries are abstractive, training our model requires creating a gold-standard set of extractive summaries. We present 
a new approach for
aligning chapter sentences with the abstractive summary sentences, incorporating weighting to ROUGE~\cite{lin-2004-rouge} and METEOR~\cite{lavie2009meteor} metrics to enable the alignment of salient words between them.  We also experiment with BERT~\cite{devlin2018bert} alignment.

We use a stable matching algorithm to select the best alignments, and show that enforcing one-to-one alignments between reference summary sentences and chapter sentences is the best alignment method of those used in earlier work. 

We obtain a dataset of summaries from five study guide websites paired with chapter text from Project Gutenberg.
Our dataset consists of 4,383 unique chapters, each of which is paired with two to five human-written summaries.

We experiment with generating summaries using our new alignment method within three models that have been developed for single document news summarization~\cite{chen2018fast,kedzie2018content,nallapati2017summarunner}. Our evaluation using automated metrics as well as a crowd-sourced pyramid evaluation shows that using the new alignment method produces significantly better results than prior work.

We also experiment with extraction at different levels of granularity, hypothesizing that extracting constituents will work better than extracting sentences, since summary sentences often combine information from several different chapter sentences. Here, our results are mixed and we offer an explanation for why this might be the case.

Our contributions include a new, challenging summarization task, experimentation that reveals potential problems with previous methods for creating extracts, and an improved method for creating gold standard extracts.
\section{Related Work}
Relatively little work has been done in summarization of novels, but early work~\cite{mihalcea-ceylan-2007-explorations} provided a dataset of novel/summary pairs drawn from CliffsNotes and GradeSaver and developed an unsupervised system based on Meade~\cite{Radev&al.01h} and TextRank~\cite{Mihalcea04TextRank} that showed promise. More recently, \citet{DBLP:conf/aaai/ZhangCO19} developed an approach for summarizing characters within a novel. We hypothesize that our  proposed task is more feasible than summarizing the full novel. 

Previous work has summarized documents using Rhetorical Structure Theory (RST)~\cite{Mann&Thompson88} to extract elementary discourse units (EDUs) for compression and more content-packed summaries~\cite{daume2002noisy,li-etal-2016-role,arumae-etal-2019-towards}. Some abstractive neural methods propose attention to focus on  phrases within a sentence to extract~\cite{gehrmann-etal-2018-bottom}. Fully abstractive methods are not yet appropriate for our task due to extensive paraphrasing and generalization.

While previous work on semantic textual similarity is relevant to the problem of finding alignments between chapter and summary text, the data available \cite{cer-etal-2017-semeval, dolan-brockett-2005-automatically} is not suitable for our domain, and the alignments generated from this data were of a poorer quality than the other methods in our paper.
\section{Data}
\label{sec:data}

We collect summary-chapter pairs from five online study guides: 
\href{http://web.archive.org/web/20190904073146/http://barronsbooknotes.com/}{BarronsBookNotes (BB)}, \href{http://www.bookwolf.com/}{BookWolf (BW)}, \href{https://www.cliffsnotes.com/}{CliffsNotes (CN)}, \href{https://www.gradesaver.com/}{GradeSaver (GS)} and \href{http://www.novelguide.com/}{NovelGuide (NG)}.\footnote{We do not have the rights to redistribute the data. The data collection scripts are available at \url{https://github.com/manestay/novel-chapter-dataset}} We select summaries from these sources for which the complete novel text can be found on Project Gutenberg. 

Our initial dataset, for summaries with two or more sources, includes 9,560 chapter/summary pairs for 4,383 chapters drawn from 79 unique books. As our analysis shows a very long tail, two rounds of filtering were applied. First, we remove reference texts with \textgreater 700 sentences, as these are too large to fit into mini-batches ($\sim$10\% of data). Second, we remove summaries with a compression ratio of \textless 2.0, as such wordy summaries often contain a lot of commentary (i.e. phrases that have no correspondence in the chapter, $\sim$5\%).

This results in 8,088 chapter/summary pairs, and we randomly assign each book to train, development and test 
splits (6,288/938/862 pairs respectively). After filtering, chapters are on average seven times longer than news articles from CNN/Dailymail (5,165 vs 761 words), and chapter summaries are eight times longer than news summaries (372 vs 46 words).

Train split statistics are given in Table~\ref{tab:dataset}. These statistics reveal the large variation in length.
Furthermore, we calculate \textit{word overlap}, the proportion of vocabulary that overlaps between the summary and chapter. 
For novels, this is 33.7\%; for CNN/DailyMail news, this is 68.7\%. This indicates  the large amount of paraphrasing in the chapter summaries in relation to the original chapter.

In Figure~\ref{fig:paraphrasing}, we show the first three sentences of a reference summary for Chapter 11, {\em The Awakening} which is paraphrased from several, non-consecutive chapter sentences shown near the bottom of the figure. We also show a portion of the summaries from two other sources which convey the same content and illustrate the extreme level of paraphrasing as well as differences in detail.  We show the full chapter and three full reference summaries in Appendix~\ref{app:example_ref}. 

\begin{table}
    \centering
    \scalebox{0.9}{
    \begin{tabular}{c c c | c}
    \hline
         \toprule
         Summary Src   &  Mean (stdev) & Median &  Total \# \\
         \midrule
         CN & 442 (369) & 347 & 1,053\\
         BB & 517 (388) & 429 & 1,000\\
         GS & 312 (311) & 230 & 1,983\\
         BW & 276 (232) & 214 & 182\\
         NG & 334 (302) & 244 & 2,070\\
         All Sources & 373 (339) & 279 & 6,288\\
         \midrule
         Chapter Text & 5,165 (3,737) & 4,122 & 6,288\\
         \bottomrule
        \hline
    \end{tabular}
    }
    \caption{\textbf{Train Split Statistics}: World count statistics with total number for summaries and chapter text.}
    \label{tab:dataset}
\end{table}

\begin{figure}[!ht]
\centering
\scalebox{0.9}{
\fbox{
\begin{minipage}{\linewidth}
{\bf GS:} In this chapter Mr. and Mrs. Pontellier participate in a battle of wills. When Mr. Pontellier gets back from the beach, he asks his wife to come inside. She tells him not to wait for her, at which point he becomes irritable and more forcefully tells her to come inside.\\
{\bf NG: } Mr. Pontellier is surprised to find Edna still outside when he returns from escorting Madame Lebrun home. ... although he asks her to come in to the house with him, she refuses, and remains outside, exercising her own will.\\\
{\bf BW:} Leonce urges Edna to go to bed, but she is still exhilarated and decides to stay outside in the hammock... \\ 
{\bf Chapter sentences:} He had walked up with Madame Lebrun and left her at the house. "Do you know it is past one o'clock? Come on," and he mounted the steps and went into their room. ``Don't wait for me,'' she answered. ``You will take cold out there,'' he said, irritably. ``What folly is this? Why don't you come in?''
\end{minipage}
}}
\caption{Portions of three reference summaries for {\em The Awakening}, Chapter 11 by Kate Chopin, along with chapter sentences they summarize.}
\label{fig:paraphrasing}
\end{figure}

\section{Alignment Experiments}
\label{section:alignments}
To train models for content selection, we need saliency labels for each chapter segment that serve as proxy extract labels, since there are no gold extracts. In news summarization, these are typically produced by aligning 
reference summaries to the best matching sentences from the news article. Here, we align the reference summary sentences with sentences from the chapter. 

We address two questions for aligning chapter and summary sentences to  generate gold standard extracts: 1) Which similarity metric works best for alignment (Section~\ref{section:methods})?  and 2) Which alignment method works best (Section~\ref{section:optimization})?

\subsection{Similarity Metrics}
\label{section:methods}
ROUGE is commonly used as a similarity metric to align the input document and the gold standard summary to produce gold extracts \cite{chen2018fast, nallapati2017summarunner, kedzie2018content}. One drawback to using ROUGE as a similarity metric is that it weights all words equally. We want to, instead, assign a higher weight for the salient words of a particular sentence.

To achieve this, we incorporate a smooth inverse frequency weighting scheme~\cite{DBLP:conf/iclr/AroraLM17} to compute word weights.
The weight of a given word is computed as follows:

\begin{equation}
\resizebox{.35\hsize}{!}{
  $W(w_i) = \frac{\alpha}{\alpha+p(w_i)}$}
\end{equation}
where $p(w_i)$ is estimated from the chapter text
and 
$\alpha$ is a smoothing parameter (here $\alpha = 1e^{-3}$).
N-gram and Longest Common Subsequence (LCS) weights are derived by summing the weights of each of the individual words in the N-gram/LCS. 
We take the average of ROUGE-{1, 2, L} using this weighting scheme as the metric for generating extracts, {\em R-wtd}, incorporating a stemmer to match morphological variants~\cite{porter1980algorithm}.

{\bf Similarity Metrics Results:} We compare R-wtd against ROUGE-L~\cite{chen2018fast} (R-L), and ROUGE-1, with stop-word removal and stemming~\cite{kedzie2018content} (R-1), for sentence alignment. 
To incorporate paraphrasing, we average METEOR~\cite{banerjee-lavie-2005-meteor} scores with ROUGE-{1,2,L} for both un-weighted (RM) and weighted scores (RM-wtd). Given the recent success of large, pre-trained language models for downstream NLP tasks, we also experiment with BERT~\cite{devlin-etal-2019-bert} to compute alignment, using cosine similarity between averaged chapter segment and summary segment vectors. We compare the generated gold extracts using R-L F1 against reference summaries, to determine a shortlist for human evaluation (to save costs). 

For the human evaluation, we ask crowd workers to measure content overlap between the generated alignments, and the reference summary, on a subset of the validation data. For each summary reference, they are shown a generated alignment and asked to indicate whether it conveys each of up to 12 summary reference sentences. An example task is shown in Appendix Figure \ref{fig:hit}. We then compute precision and recall based on the number of summary sentences conveyed in the extract.

Table~\ref{tab:human_metric} shows that humans prefer alignments generated using R-wtd by a significant margin.\footnote{We suspect incorporating METEOR by averaging didn't work because the scale is different from ROUGE scores.}
Sample alignments generated by R-wtd in comparison to the baseline are shown in Figure~\ref{fig:align}.

\begin{table}[ht]
\centering
\setlength{\tabcolsep}{3pt}
\scalebox{.9}{
\begin{tabular}[t]{lcccccc}
\toprule
Method & RM & R-wtd & RM-wtd & R-1 & R-L & BERT \\
\midrule
R-L F1 & \textbf{41.2} & \textbf{40.6} & 39.3 & 37.1 & 35.1 & 35.4 \\
H-F1 & 33.7 &  \textbf{44.8}& 38.8 & -- & -- & --\\
\bottomrule
\end{tabular}
}
\caption{ROUGE-L F1, and crowd-sourced F1 scores (H-F1) for content overlap.}
\label{tab:human_metric}
\end{table}%

\subsection{Alignment Methods}
\label{section:optimization}
Some previous work in news summarization has focused on iteratively picking the best article sentence with respect to the summary, in order to get the gold extracts~\cite{nallapati2017summarunner, kedzie2018content}, using ROUGE between the set of selected sentences and the target summary.
In contrast, others have focused on picking the best article sentence with respect to each sentence in the summary~\cite{chen2018fast}. We investigate which approach yields better alignments. We refer to the former method as summary-level alignment and the latter method as sentence-level alignment. 

For sentence-level alignment, we note that the problem of finding optimal alignments is similar to a stable matching problem. We wish to find a set of alignments such that there exists no chapter segment $a$ and summary segment $x$ where both $a$ and $x$ would prefer to be aligned with each other over their current alignment match. We compute alignments based on the Gale-Shapley algorithm~\shortcite{gale1962college} for stable matching and compare it with the greedy approach from prior work~\cite{chen2018fast}.

For summary-level alignment~\cite{nallapati2017summarunner, kedzie2018content}, we compare two variants: selecting sentences until we reach the reference word count (WL summary), and selecting sentences until the ROUGE score no longer increases (WS summary).

Crowd-sourced evaluation results (Table~\ref{tab:human_optimization}) show that sentence-level stable matching is significantly better. We use this in the remainder of this work.
These differences in alignments affect earlier claims about the performance of summarization systems, as they were not measured, yet have a significant impact.\footnote{Bold text indicates statistical significance with $p<0.05$.}

\begin{table}[ht]
\centering
\scalebox{.9}{
\begin{tabular}[t]{lccc}
\toprule
Method&P&R&F1\\
\midrule
Greedy Sent&48.4&48.7&48.5\\
Stable Sent&\textbf{52.8}&\textbf{52.6}&\textbf{52.7}\\
WL summary&34.5&36.6&36.7\\
WS summary&42.7&36.6&38.0\\
\bottomrule
\end{tabular}
}
\caption{Crowd sourced evaluation on content overlap for summary vs. sentence level on validation set.}
\label{tab:human_optimization}
\end{table}%

\begin{figure}[!ht]
\centering
\scalebox{0.9}{
\fbox{
\begin{minipage}{\linewidth}
{\bf Ref summary:} He says he will, as soon as he has finished his last cigar.\\
{\bf R-L greedy:} ``You will take cold out there,'' he said, irritably.\\
{\bf R-L stable:} He drew up the rocker, hoisted his slippered feet on the rail, and proceeded to smoke a cigar.\\
{\bf R-wtd stable:} ``Just as soon as I have finished my cigar.''
\end{minipage}
}}
\caption{A reference summary sentence and its alignments. R-L greedy and R-L stable are incorrect because they weight words equally (e.g. said, cigar, `.').}
\label{fig:align}
\end{figure}
\section{Summarization Experiments}

In order to assess how alignments impact summarization,  
we train three extractive systems --  hierarchical CNN-LSTM extractor \cite{chen2018fast} (CB), seq2seq with attention \cite{kedzie2018content} (K), and RNN \cite{nallapati2017summarunner} (N).
The target word length of generated summaries is based on the average summary length of similarly long chapters from the training set.\footnote{We do so by binning chapters into 10 quantiles by length.}

We also experiment with aligning and extracting at the constituent level,\footnote{Prior work has used EDUs, but automated parsers such as \cite{ji2014representation} perform poorly in this domain.} given our observation during data analysis that summary sentences are often drawn from two different chapter sentences. 
We create syntactic constituents by taking sub-trees from constituent parse trees for each sentence~\cite{manning-EtAl:2014:P14-5} rooted with \texttt{S}-tags. To ensure that constituents are long enough to be meaningful, we take the longest \texttt{S}-tag when one \texttt{S}-tag is embedded within others (see Appendix \ref{app:constituent}).

Summary quality is evaluated on F1 scores for R-\{1,2,L\}, and METEOR. Each chapter has 2-5 reference summaries and we evaluate the generated summary against all the reference summaries. Part of a generated summary of extracted constituents for Chapter 11, {\em The Awakening}, is shown in Figure~\ref{fig:output}. The full generated summaries for this chapter (both extracted constituents and extracted sentences) are shown in Appendix~\ref{app:example_ref}.

\begin{figure}[!ht]
\scalebox{.9}{\fbox{
\begin{minipage}{\linewidth}
{\bf Generated Summary: }  \textbar \textcolor{teal} {I thought I should find you in bed , ''} \textbar \textbar \textcolor{teal} { said her husband }, \textbar \textcolor{teal} {when he discovered her} \textbar lying there . \textbar \textcolor{teal} {He had walked up with Madame Lebrun and left her at the house .} \textbar \textbar \textcolor{teal} {She heard him moving about the room ;} \textbar every sound indicating impatience and irritation .\textbar 
\end{minipage}
}}
\caption{System generated summary, extracted constituents in \textcolor{teal}{teal}, and separated by \textbar. }
    \label{fig:output}
\end{figure}

\subsection{Results}

We compare our method for generating extractive targets (ROUGE weighted, with stable matching at the sentence level) against the baseline method for generating extractive targets for each of the systems.
Table \ref{tab:results_trained} shows three rows for each summarization system: using the original target summary labels, and using either constituent or sentence  segments. We see our proposed alignment method performs significantly better for all models. ROUGE-L in particular increases 10\% to 18\% relatively over the baselines. Moreover, it would seem at first glance that the K and N baseline models perform better than the CB baseline, however this difference has nothing to do with the architecture choice. When we use our extractive targets, all three models perform similarly, suggesting that the differences are mainly due to small, but important, differences in their methods for generating extractive targets.

\begin{table}[t]
\centering
\setlength{\tabcolsep}{3pt}
\scalebox{0.9}{
\begin{tabular}{l l l l l l l}
\toprule
Model & Seg & Method & R-1 & R-2 & R-L & METEOR \\
\midrule
\multirow{3}{1.3em}{CB} & sent & baseline & 33.1 & 5.5 & 30.0 & 13.9 \\
& sent & R-wtd & \textbf{35.8} & \textbf{6.9} & 33.4 & \textbf{15.2} \\
& const & R-wtd & \textbf{36.2} & \textbf{6.9} & \textbf{35.4} & \textbf{15.2} \\
\midrule
\multirow{3}{1em}{K} & sent & baseline & 34.3 & 6.4 & 31.6 & 14.6 \\
& sent & R-wtd & \textbf{35.6} & \textbf{6.9} & 33.2 & \textbf{15.0} \\
& const & R-wtd &  \textbf{36.2} & \textbf{6.9} & \textbf{35.2} & \textbf{15.1} \\
\midrule
\multirow{3}{1em}{N} & sent & baseline & 34.6 & 6.4 & 31.9 & 14.6 \\
& sent & R-wtd & \textbf{35.7} & \textbf{7.0} & 33.3 & \textbf{15.1} \\
& const & R-wtd & \textbf{35.9} & \textbf{7.0} & \textbf{35.2} & \textbf{15.0} \\
\bottomrule
\end{tabular}
}
\caption{ROUGE-F1, METEOR for generated summaries. "Baseline" is the method used for that model.}
\label{tab:results_trained}
\end{table}

{\bf Human Evaluation:}
\label{section:human}
Given questions about the reliability of ROUGE~\cite{novikova-etal-2017-need, chaganty-etal-2018-price}, we perform  human evaluation to assess which system is  best at content selection. 
We use  a lightweight, sampling based approach for pyramid analysis that relies on crowd-sourcing, proposed by \newcite{shapira-etal-2019-crowdsourcing}, and correlates well with the original pyramid method~\cite{Nenkova:2007:PMI:1233912.1233913}. We ask the crowd workers to indicate which of the sampled  reference summary content units are conveyed in the generated summary.\footnote{See the screen shot in Appendix \ref{app:scu_eval}}

We evaluated our best system + alignment on extraction of sentences and of constituents (CB R-wtd), along with a baseline system (CB K-align),\footnote{We use the best baseline alignment, \newcite{kedzie2018content} with the CB model to keep model choice consistent.} using the crowd-sourced pyramid evaluation method. To produce readable summaries for extracted constituents, each extracted constituent is included along with the context of the containing sentence (black text in Figure~\ref{fig:output}). We find that CB Sent R-wtd has significantly higher content overlap with reference summaries in Table \ref{tab:pyramid}.

\begin{table}[ht]
\centering
\begin{tabular}[t]{lc}
\toprule
System&Pyramid Score\\
\midrule
CB K-align&17.9\\
CB Sent R-wtd&\textbf{18.9}\\
CB Const R-wtd&18.1\\
\bottomrule
\end{tabular}
\caption{Crowd-sourced Pyramid Evaluation.}
\label{tab:pyramid}
\end{table}
\section{Discussion and Conclusion}

We present a new challenging task for summarization of novel chapters. We show that \textit{sentence-level, stable-matched} alignment is better than the summary-level alignment used in previous work and our proposed \textit{R-wtd} method for creating gold extracts is shown to be better than other similarity metrics. The resulting system is the first step towards addressing this task. 

While both human evaluation and automated metrics concur that summaries produced with our new alignment approach outperform previous approaches, they contradict on the question of whether extraction is better at the constituent or the sentence level. We hypothesize that because we use ROUGE to score summaries of extracted constituents without context, the selected content is packed into the word budget; there is no potentially irrelevant context to count against the system. In contrast, we do include sentence context in the pyramid evaluation in order to make the summaries readable for humans and thus, fewer constituents make it into the generated summary for the human evaluation. This could account for the increased score on automated metrics.

It is also possible that smaller constituents can be matched to phrases within the summary with metrics such as ROUGE, when they actually should not have counted. In future work, we plan to experiment more with this, examining how we can combine constituents to make fluent sentences without including potentially irrelevant context.

We would also like to further experiment with abstractive summarization to re-examine whether large, pre-trained language models~\cite{liu-lapata-2019-text} can be improved for our domain.
We suspect these models are problematic for our documents because they are, on average, an order of magnitude larger than what was used for pre-training the language model (512 tokens). Another issue is that the pre-trained language models are very large and take up a substantial amount of GPU memory, which limits how long the input document can be. While truncation of a document may not hurt performance in the news domain due to the heavy lede bias, in our domain, truncation can hurt the performance of the summarizer.

\bibliography{novel,anthology}

\begin{thebibliography}{32}
\expandafter\ifx\csname natexlab\endcsname\relax\def\natexlab#1{#1}\fi

\bibitem[{Arora et~al.(2017)Arora, Liang, and Ma}]{DBLP:conf/iclr/AroraLM17}
Sanjeev Arora, Yingyu Liang, and Tengyu Ma. 2017.
\newblock \href {https://openreview.net/forum?id=SyK00v5xx} {A simple but
  tough-to-beat baseline for sentence embeddings}.
\newblock In \emph{5th International Conference on Learning Representations,
  {ICLR} 2017, Toulon, France, April 24-26, 2017, Conference Track
  Proceedings}.

\bibitem[{Arumae et~al.(2019)Arumae, Bhatia, and
  Liu}]{arumae-etal-2019-towards}
Kristjan Arumae, Parminder Bhatia, and Fei Liu. 2019.
\newblock \href {https://doi.org/10.18653/v1/D19-5408} {Towards annotating and
  creating summary highlights at sub-sentence level}.
\newblock In \emph{Proceedings of the 2nd Workshop on New Frontiers in
  Summarization}, pages 64--69, Hong Kong, China. Association for Computational
  Linguistics.

\bibitem[{Banerjee and Lavie(2005)}]{banerjee-lavie-2005-meteor}
Satanjeev Banerjee and Alon Lavie. 2005.
\newblock \href {https://www.aclweb.org/anthology/W05-0909} {{METEOR}: An
  automatic metric for {MT} evaluation with improved correlation with human
  judgments}.
\newblock In \emph{Proceedings of the {ACL} Workshop on Intrinsic and Extrinsic
  Evaluation Measures for Machine Translation and/or Summarization}, pages
  65--72, Ann Arbor, Michigan. Association for Computational Linguistics.

\bibitem[{Cer et~al.(2017)Cer, Diab, Agirre, Lopez-Gazpio, and
  Specia}]{cer-etal-2017-semeval}
Daniel Cer, Mona Diab, Eneko Agirre, I{\~n}igo Lopez-Gazpio, and Lucia Specia.
  2017.
\newblock \href {https://doi.org/10.18653/v1/S17-2001} {{S}em{E}val-2017 task
  1: Semantic textual similarity multilingual and crosslingual focused
  evaluation}.
\newblock In \emph{Proceedings of the 11th International Workshop on Semantic
  Evaluation ({S}em{E}val-2017)}, pages 1--14, Vancouver, Canada. Association
  for Computational Linguistics.

\bibitem[{Chaganty et~al.(2018)Chaganty, Mussmann, and
  Liang}]{chaganty-etal-2018-price}
Arun Chaganty, Stephen Mussmann, and Percy Liang. 2018.
\newblock \href {https://doi.org/10.18653/v1/P18-1060} {The price of debiasing
  automatic metrics in natural language evalaution}.
\newblock In \emph{Proceedings of the 56th Annual Meeting of the Association
  for Computational Linguistics (Volume 1: Long Papers)}, pages 643--653,
  Melbourne, Australia. Association for Computational Linguistics.

\bibitem[{Chen and Bansal(2018)}]{chen2018fast}
Yen-Chun Chen and Mohit Bansal. 2018.
\newblock Fast abstractive summarization with reinforce-selected sentence
  rewriting.
\newblock In \emph{Proceedings of the 56th Annual Meeting of the Association
  for Computational Linguistics (Volume 1: Long Papers)}, pages 675--686.

\bibitem[{Cheng and Lapata(2016)}]{cheng-lapata-2016-neural}
Jianpeng Cheng and Mirella Lapata. 2016.
\newblock \href {https://doi.org/10.18653/v1/P16-1046} {Neural summarization by
  extracting sentences and words}.
\newblock In \emph{Proceedings of the 54th Annual Meeting of the Association
  for Computational Linguistics (Volume 1: Long Papers)}, pages 484--494,
  Berlin, Germany. Association for Computational Linguistics.

\bibitem[{Daum{\'e}~III and Marcu(2002)}]{daume2002noisy}
Hal Daum{\'e}~III and Daniel Marcu. 2002.
\newblock A noisy-channel model for document compression.
\newblock In \emph{Proceedings of the 40th Annual Meeting on Association for
  Computational Linguistics}, pages 449--456. Association for Computational
  Linguistics.

\bibitem[{Devlin et~al.(2018)Devlin, Chang, Lee, and
  Toutanova}]{devlin2018bert}
Jacob Devlin, Ming-Wei Chang, Kenton Lee, and Kristina Toutanova. 2018.
\newblock \href {http://arxiv.org/abs/1810.04805} {Bert: Pre-training of deep
  bidirectional transformers for language understanding}.

\bibitem[{Devlin et~al.(2019)Devlin, Chang, Lee, and
  Toutanova}]{devlin-etal-2019-bert}
Jacob Devlin, Ming-Wei Chang, Kenton Lee, and Kristina Toutanova. 2019.
\newblock \href {https://doi.org/10.18653/v1/N19-1423} {{BERT}: Pre-training of
  deep bidirectional transformers for language understanding}.
\newblock In \emph{Proceedings of the 2019 Conference of the North {A}merican
  Chapter of the Association for Computational Linguistics: Human Language
  Technologies, Volume 1 (Long and Short Papers)}, pages 4171--4186,
  Minneapolis, Minnesota. Association for Computational Linguistics.

\bibitem[{Dolan and Brockett(2005)}]{dolan-brockett-2005-automatically}
William~B. Dolan and Chris Brockett. 2005.
\newblock \href {https://www.aclweb.org/anthology/I05-5002} {Automatically
  constructing a corpus of sentential paraphrases}.
\newblock In \emph{Proceedings of the Third International Workshop on
  Paraphrasing ({IWP}2005)}.

\bibitem[{Gale and Shapley(1962)}]{gale1962college}
David Gale and Lloyd~S Shapley. 1962.
\newblock College admissions and the stability of marriage.
\newblock \emph{The American Mathematical Monthly}, 69(1):9--15.

\bibitem[{Gehrmann et~al.(2018)Gehrmann, Deng, and
  Rush}]{gehrmann-etal-2018-bottom}
Sebastian Gehrmann, Yuntian Deng, and Alexander Rush. 2018.
\newblock \href {https://doi.org/10.18653/v1/D18-1443} {Bottom-up abstractive
  summarization}.
\newblock In \emph{Proceedings of the 2018 Conference on Empirical Methods in
  Natural Language Processing}, pages 4098--4109, Brussels, Belgium.
  Association for Computational Linguistics.

\bibitem[{Grusky et~al.(2018)Grusky, Naaman, and
  Artzi}]{grusky-etal-2018-newsroom}
Max Grusky, Mor Naaman, and Yoav Artzi. 2018.
\newblock \href {https://doi.org/10.18653/v1/N18-1065} {{N}ewsroom: A dataset
  of 1.3 million summaries with diverse extractive strategies}.
\newblock In \emph{Proceedings of the 2018 Conference of the North {A}merican
  Chapter of the Association for Computational Linguistics: Human Language
  Technologies, Volume 1 (Long Papers)}, pages 708--719, New Orleans,
  Louisiana. Association for Computational Linguistics.

\bibitem[{Ji and Eisenstein(2014)}]{ji2014representation}
Yangfeng Ji and Jacob Eisenstein. 2014.
\newblock Representation learning for text-level discourse parsing.
\newblock In \emph{Proceedings of the 52nd Annual Meeting of the Association
  for Computational Linguistics (Volume 1: Long Papers)}, volume~1, pages
  13--24.

\bibitem[{Kedzie et~al.(2018)Kedzie, McKeown, and
  Daume~III}]{kedzie2018content}
Chris Kedzie, Kathleen McKeown, and Hal Daume~III. 2018.
\newblock Content selection in deep learning models of summarization.
\newblock In \emph{Proceedings of the 2018 Conference on Empirical Methods in
  Natural Language Processing}, pages 1818--1828.

\bibitem[{Lavie and Denkowski(2009)}]{lavie2009meteor}
Alon Lavie and Michael~J Denkowski. 2009.
\newblock The meteor metric for automatic evaluation of machine translation.
\newblock \emph{Machine translation}, 23(2-3):105--115.

\bibitem[{Li et~al.(2016)Li, Thadani, and Stent}]{li-etal-2016-role}
Junyi~Jessy Li, Kapil Thadani, and Amanda Stent. 2016.
\newblock \href {https://doi.org/10.18653/v1/W16-3617} {The role of discourse
  units in near-extractive summarization}.
\newblock In \emph{Proceedings of the 17th Annual Meeting of the Special
  Interest Group on Discourse and Dialogue}, pages 137--147, Los Angeles.
  Association for Computational Linguistics.

\bibitem[{Lin(2004)}]{lin-2004-rouge}
Chin-Yew Lin. 2004.
\newblock \href {https://www.aclweb.org/anthology/W04-1013} {{ROUGE}: A package
  for automatic evaluation of summaries}.
\newblock In \emph{Text Summarization Branches Out}, pages 74--81, Barcelona,
  Spain. Association for Computational Linguistics.

\bibitem[{Liu and Lapata(2019)}]{liu-lapata-2019-text}
Yang Liu and Mirella Lapata. 2019.
\newblock \href {https://doi.org/10.18653/v1/D19-1387} {Text summarization with
  pretrained encoders}.
\newblock In \emph{Proceedings of the 2019 Conference on Empirical Methods in
  Natural Language Processing and the 9th International Joint Conference on
  Natural Language Processing (EMNLP-IJCNLP)}, pages 3728--3738, Hong Kong,
  China. Association for Computational Linguistics.

\bibitem[{Mann and Thompson(1988)}]{Mann&Thompson88}
William~C. Mann and Sandra~A. Thompson. 1988.
\newblock \href {https://doi.org/doi:10.1515/text.1.1988.8.3.243} {Rhetorical
  structure theory: toward a functional theory of text organization}.
\newblock \emph{Text: Interdisciplinary Journal for the Study of Discourse},
  8(3):243–281.

\bibitem[{Manning et~al.(2014)Manning, Surdeanu, Bauer, Finkel, Bethard, and
  McClosky}]{manning-EtAl:2014:P14-5}
Christopher~D. Manning, Mihai Surdeanu, John Bauer, Jenn\~y Finkel, Steven~J.
  Bethard, and David McClosky. 2014.
\newblock \href {http://www.aclweb.org/anthology/P/P14/P14-5010} {The
  {Stanford} {CoreNLP} natural language processing toolkit}.
\newblock In \emph{Association for Computational Linguistics (ACL) System
  Demonstrations}, pages 55--60.

\bibitem[{Mihalcea and Tarau(2004)}]{Mihalcea04TextRank}
R.~Mihalcea and P.~Tarau. 2004.
\newblock {TextRank}: Bringing order into texts.
\newblock In \emph{Proceedings of {EMNLP-04}and the 2004 Conference on
  Empirical Methods in Natural Language Processing}.

\bibitem[{Mihalcea and Ceylan(2007)}]{mihalcea-ceylan-2007-explorations}
Rada Mihalcea and Hakan Ceylan. 2007.
\newblock \href {https://www.aclweb.org/anthology/D07-1040} {Explorations in
  automatic book summarization}.
\newblock In \emph{Proceedings of the 2007 Joint Conference on Empirical
  Methods in Natural Language Processing and Computational Natural Language
  Learning ({EMNLP}-{C}o{NLL})}, pages 380--389, Prague, Czech Republic.
  Association for Computational Linguistics.

\bibitem[{Nallapati et~al.(2017)Nallapati, Zhai, and
  Zhou}]{nallapati2017summarunner}
Ramesh Nallapati, Feifei Zhai, and Bowen Zhou. 2017.
\newblock Summarunner: A recurrent neural network based sequence model for
  extractive summarization of documents.
\newblock In \emph{Thirty-First AAAI Conference on Artificial Intelligence}.

\bibitem[{Nenkova et~al.(2007)Nenkova, Passonneau, and
  McKeown}]{Nenkova:2007:PMI:1233912.1233913}
Ani Nenkova, Rebecca Passonneau, and Kathleen McKeown. 2007.
\newblock \href {https://doi.org/10.1145/1233912.1233913} {The pyramid method:
  Incorporating human content selection variation in summarization evaluation}.
\newblock \emph{ACM Trans. Speech Lang. Process.}, 4(2).

\bibitem[{Novikova et~al.(2017)Novikova, Du{\v{s}}ek, Cercas~Curry, and
  Rieser}]{novikova-etal-2017-need}
Jekaterina Novikova, Ond{\v{r}}ej Du{\v{s}}ek, Amanda Cercas~Curry, and Verena
  Rieser. 2017.
\newblock \href {https://doi.org/10.18653/v1/D17-1238} {Why we need new
  evaluation metrics for {NLG}}.
\newblock In \emph{Proceedings of the 2017 Conference on Empirical Methods in
  Natural Language Processing}, pages 2241--2252, Copenhagen, Denmark.
  Association for Computational Linguistics.

\bibitem[{Paulus et~al.(2017)Paulus, Xiong, and Socher}]{paulus2017deep}
Romain Paulus, Caiming Xiong, and Richard Socher. 2017.
\newblock A deep reinforced model for abstractive summarization.
\newblock \emph{arXiv preprint arXiv:1705.04304}.

\bibitem[{Porter(1980)}]{porter1980algorithm}
Martin~F Porter. 1980.
\newblock An algorithm for suffix stripping.
\newblock \emph{Program}, 14(3):130--137.

\bibitem[{Radev et~al.(2001)Radev, Blair-Goldensohn, and Zhang}]{Radev&al.01h}
Dragomir Radev, Sasha Blair-Goldensohn, and Zhu Zhang. 2001.
\newblock Experiments in single and multi-document summarization using {MEAD}.
\newblock In \emph{First Document Understanding Conference}, New Orleans, LA.

\bibitem[{Shapira et~al.(2019)Shapira, Gabay, Gao, Ronen, Pasunuru, Bansal,
  Amsterdamer, and Dagan}]{shapira-etal-2019-crowdsourcing}
Ori Shapira, David Gabay, Yang Gao, Hadar Ronen, Ramakanth Pasunuru, Mohit
  Bansal, Yael Amsterdamer, and Ido Dagan. 2019.
\newblock \href {https://doi.org/10.18653/v1/N19-1072} {Crowdsourcing
  lightweight pyramids for manual summary evaluation}.
\newblock In \emph{Proceedings of the 2019 Conference of the North {A}merican
  Chapter of the Association for Computational Linguistics: Human Language
  Technologies, Volume 1 (Long and Short Papers)}, pages 682--687, Minneapolis,
  Minnesota. Association for Computational Linguistics.

\bibitem[{Zhang et~al.(2019)Zhang, Cheung, and Oren}]{DBLP:conf/aaai/ZhangCO19}
Weiwei Zhang, Jackie Chi~Kit Cheung, and Joel Oren. 2019.
\newblock \href {https://doi.org/10.1609/aaai.v33i01.33017476} {Generating
  character descriptions for automatic summarization of fiction}.
\newblock In \emph{The Thirty-Third {AAAI} Conference on Artificial
  Intelligence, {AAAI} 2019, The Thirty-First Innovative Applications of
  Artificial Intelligence Conference, {IAAI} 2019, The Ninth {AAAI} Symposium
  on Educational Advances in Artificial Intelligence, {EAAI} 2019, Honolulu,
  Hawaii, USA, January 27 - February 1, 2019}, pages 7476--7483.

\end{thebibliography}
\bibliographystyle{acl_natbib}

\clearpage
\newpage
\appendix
\section{Appendix}

\subsection{Acknowledgments}
We would like to thank Spandana Gella for her contributions to the project. 
We would like to thank Jonathan Steuck, Alessandra Brusadin, and the rest of the AWS AI Data team for their invaluable feedback in the data annotation process.
We would finally like to thank Christopher Hidey, Christopher Kedzie, Emily Allaway, Esin Durmus, Fei-Tzin Lee, Feng Nan, Miguel Ballesteros, Ramesh Nallapati, and the anonymous reviewers for their valuable feedback on this paper.

\subsection{Example Chapter and Summaries}
\label{app:example_ref}

We show the full text of Chapter 11, \textit{The Awakening} by Kate Chopin in Figure \ref{fig:chapter_text}. We show three reference summaries in Figure \ref{fig:ref_summ}, and two generated summaries using our best alignment method in Figure \ref{fig:gen_summ}. While there are differences in length and level of detail, there are also clearly similarities in covered content.

\begin{figure*}[!ht]
\fbox{
\begin{minipage}{.95\linewidth}
 ``What are you doing out here, Edna? I thought I should find you in bed,'' said her husband, when he discovered her lying there. He had walked up with Madame Lebrun and left her at the house. His wife did not reply.

``Are you asleep?'' he asked, bending down close to look at her.

``No.'' Her eyes gleamed bright and intense, with no sleepy shadows, as they looked into his.

``Do you know it is past one o'clock? Come on,'' and he mounted the steps and went into their room.

``Edna!'' called Mr. Pontellier from within, after a few moments had gone by.

``Don't wait for me,'' she answered. He thrust his head through the door.

``You will take cold out there,'' he said, irritably. ``What folly is this? Why don't you come in?''

``It isn't cold; I have my shawl.''

``The mosquitoes will devour you.''

``There are no mosquitoes.''

She heard him moving about the room; every sound indicating impatience and irritation. Another time she would have gone in at his request. She would, through habit, have yielded to his desire; not with any sense of submission or obedience to his compelling wishes, but unthinkingly, as we walk, move, sit, stand, go through the daily treadmill of the life which has been portioned out to us.

``Edna, dear, are you not coming in soon?'' he asked again, this time fondly, with a note of entreaty.

``No; I am going to stay out here.''

``This is more than folly,'' he blurted out. ``I can't permit you to stay out there all night. You must come in the house instantly.''

With a writhing motion she settled herself more securely in the hammock. She perceived that her will had blazed up, stubborn and resistant. She could not at that moment have done other than denied and resisted. She wondered if her husband had ever spoken to her like that before, and if she had submitted to his command. Of course she had; she remembered that she had. But she could not realize why or how she should have yielded, feeling as she then did.

``Leonce, go to bed,'' she said, ``I mean to stay out here. I don't wish to go in, and I don't intend to. Don't speak to me like that again; I shall not answer you.''

Mr. Pontellier had prepared for bed, but he slipped on an extra garment. He opened a bottle of wine, of which he kept a small and select supply in a buffet of his own. He drank a glass of the wine and went out on the gallery and offered a glass to his wife. She did not wish any. He drew up the rocker, hoisted his slippered feet on the rail, and proceeded to smoke a cigar. He smoked two cigars; then he went inside and drank another glass of wine. Mrs. Pontellier again declined to accept a glass when it was offered to her. Mr. Pontellier once more seated himself with elevated feet, and after a reasonable interval of time smoked some more cigars.

Edna began to feel like one who awakens gradually out of a dream, a delicious, grotesque, impossible dream, to feel again the realities pressing into her soul. The physical need for sleep began to overtake her; the exuberance which had sustained and exalted her spirit left her helpless and yielding to the conditions which crowded her in.

The stillest hour of the night had come, the hour before dawn, when the world seems to hold its breath. The moon hung low, and had turned from silver to copper in the sleeping sky. The old owl no longer hooted, and the water-oaks had ceased to moan as they bent their heads.

Edna arose, cramped from lying so long and still in the hammock. She tottered up the steps, clutching feebly at the post before passing into the house.

``Are you coming in, Leonce?'' she asked, turning her face toward her husband.

``Yes, dear,'' he answered, with a glance following a misty puff of smoke. ``Just as soon as I have finished my cigar.''
\end{minipage}
}
\caption{Full chapter text. Note that this is short at 847 words, as the median chapter length is 3168 words.}
\label{fig:chapter_text}
\end{figure*}

\begin{figure*}[!ht]
\fbox{
\begin{minipage}{.95\linewidth}
{\bf BookWolf summary:} L{\'e}once urges Edna to go to bed, but she is still exhilarated and decides to stay outside in the hammock. L{\'e}once stays up with her and smokes his cigars. Edna feels defiant towards her husband and resents his control over her life. Eventually tiredness overcomes Edna and she goes to bed.\\

{\bf GradeSaver summary: }In this chapter Mr. and Mrs. Pontellier participate in a battle of wills. When Mr. Pontellier gets back from the beach, he asks his wife to come inside. She tells him not to wait for her, at which point he becomes irritable and more forcefully tells her to come inside. Mrs. Pontellier resolves not to go in and thinks about how, on another occasion, she would have just done what her husband asked, simply because of inertia. Feeling stubborn and strong, she realizes that she had never taken such a stand against her husband before.\\
Mr. Pontellier then decides to join her outside. He drinks glasses of wine and smokes a number of cigars. After awhile, Mrs. Pontellier feels like she is being awakened from a dream and realizes that she is quite fatigued. It is almost dawn. Finally getting up from the hammock, Mrs. Pontellier asks her husband if he's going to join her. He replies that he will, after he finishes his cigar.\\

{\bf NovelGuide summary: }{Mr. Pontellier is surprised to find Edna still outside when he returns from escorting Madame Lebrun home. In a small but no doubt significant exchange-considering the events of the evening, and the novel's title-her distant and unperceiving husband asks her, "Are you asleep?" Edna, with eyes "bright and intense," definitively replies, "No." Although he asks her to come in to the house with him, she refuses, and remains outside, exercising her own will. As if trying to outlast his wife, Mr. Pontellier smokes cigar after cigar next to her. Gradually, Edna succumbs to her need for sleep. She feels "like one who awakens gradually out of a . . . delicious, grotesque, impossible dream . . . ." As described in Chapter VII, then, Edna is once again undergoing what might be called a "negative" "awakening"-an "awakening" to the realities of her present life-as opposed to the "positive" awakening to new possibilities and her own self-direction, to which the nighttime swim began to expose her. As if to underscore her failure to "awaken" to herself, the chapter ends with a scene of tables being turned: as Edna goes in, she asks her husband if he will be joining her. He says he will, as soon as he has finished his last cigar. While the narrator does not record Mr. Pontellier's tone of voice, the comments seem almost scornful, mockingly echoing Edna's earlier self-assertion.}
\end{minipage}
}
\caption{Two reference summaries.}
\label{fig:ref_summ}
\end{figure*}
\begin{figure*}[!ht]
\fbox{
\begin{minipage}{.95\linewidth}
{\bf Constituent R-wtd:}  \textbar \textcolor{teal}{I thought I should find you in bed , ''} \textbar \textbar \textcolor{teal} { said her husband }, \textbar \textcolor{teal} {when he discovered her} \textbar lying there . \textbar \textcolor{teal} {He had walked up with Madame Lebrun and left her at the house .} \textbar \textbar \textcolor{teal} {She heard him moving about the room ;} \textbar every sound indicating impatience and irritation . \textbar \textcolor{teal} {`` This is more than folly , ''} \textbar  he blurted out . ` I ca n't  \textbar \textcolor{teal} {permit you to stay out there all night }.\textbar \textbar \textcolor{teal} {But she could not realize} \textbar why or how she should have yielded , feeling as she then did . \textbar \textcolor{teal} {He smoked two cigars ;} \textbar then he went inside and drank another glass of wine . She tottered up the steps , \textbar \textcolor{teal} {clutching feebly at the post before passing into the house .} \textbar she asked , \textbar \textcolor{teal}{turning her face toward her husband .} \textbar \\

\textbf{Sentence R-wtd:} \textcolor{teal}{
\textcolor{black}{\textbar} I thought I should find you in bed , '' said her husband , when he discovered her lying there . \textcolor{black}{\textbar} 
\textcolor{black}{\textbar} He had walked up with Madame Lebrun and left her at the house . \textcolor{black}{\textbar} 
\textcolor{black}{\textbar} His wife did not reply . \textcolor{black}{\textbar} 
\textcolor{black}{\textbar} `` This is more than folly , '' he blurted out . \textcolor{black}{\textbar} 
\textcolor{black}{\textbar} You must come in the house instantly . '' \textcolor{black}{\textbar} 
\textcolor{black}{\textbar} Edna began to feel like one who awakens gradually out of a dream , a delicious , grotesque , impossible dream , to feel again the realities pressing into her soul . \textcolor{black}{\textbar} 
\textcolor{black}{\textbar} She tottered up the steps , clutching feebly at the post before passing into the house . \textcolor{black}{\textbar} 
\textcolor{black}{\textbar} she asked , turning her face toward her husband . \textcolor{black}{\textbar} 
\textcolor{black}{\textbar} `` Just as soon as I have finished my cigar . '' \textcolor{black}{\textbar} }

\end{minipage}
}
\caption{Two generated summaries. Extracted segments are highlighted in \textcolor{teal}{teal}, and delineated with \textbar. Constituents are presented with context, whereas sentences extract all text.}
\label{fig:gen_summ}
\end{figure*}

\subsection{Target Word Length for Summaries}
\label{app:bins}
The target word length for generated summaries is a function of the input chapter word count ($wc_{chapter}$).

We divide the train set into 10 quantiles, and in each quantile (or bin), associate it to the mean \textit{compression ratio} ($CR$):
\begin{equation}
CR = \frac{wc_{chapter}}{wc_{ref\_summ}}
\end{equation}
\begin{equation}
CR_{quantile} = {\frac {1}{n}}\sum _{i=1}^{n}{CR_i}
\end{equation}
Where $wc_{ref_summ}$ is the word count of the reference summary, and $CR_i$ is the compression ratio of the i-th quantile item.

The target word length for the generated summary ($wc_{gen\_summ}$) is given by:
\begin{equation}
wc_{gen\_summ} = 
\frac{1}{CR_{quantile}} * wc_{chapter}
\end{equation}

Generated summaries are created by extracting segments with the highest model probability until this budget is reached (without truncation). Oracle summaries also use this target word length, but may be shorter if the original summary had few segments (as we extract one chapter segment for each summary segment).

\begin{table}[h!]
    \centering
    \begin{tabular}{cccc}
    \toprule
         Quantile &  Min wc & Max wc & CR\\
         \midrule
         1 & 44 & 1,232 & 6.67 \\
         2 & 1,233 & 1,711 & 9.09\\
         3 & 1,712 & 2,174 & 9.09\\
         4 & 2,175 & 2,758 & 10.00\\
         5 & 2,579 & 3,361 & 11.11\\
         6 & 3,362 & 4,165 & 12.5\\
         7 & 4,166 & 5,374 & 14.29\\
         8 & 5,375 & 7,762 & 14.29\\
         9 & 7,763 & 13,028 & 16.67\\
         10 & 13,029 & 70,436 & 20\\
         \bottomrule
    \end{tabular}
    \caption{\textbf{Quantiles}: For each quantile (bin), we show its max and min word words, and its compression ratio.}
    \label{tab:bins}
\end{table}

\subsection{SCU Evaluation Task Setup}
\label{app:scu_eval}
To obtain the \textit{distractors}, we sample 2 SCUs from different chapters from the same book. We insert one of them, the positive distractor, into the generated summary, as well as into the list of statements, so it will always be correct. We insert the other, the negative distractor, only into the list of statements, so it will always be incorrect.

\newcommand{\var}{\texttt}
\newcommand{\ttcmd}[1]{\textbf{\upshape{#1}}}
\newcommand\mycommfont[1]{\footnotesize\ttfamily\textcolor{teal}{#1}}
\SetCommentSty{mycommfont}

\subsection{Constituent Extraction algorithm}
\label{app:constituent}
Algorithm~\ref{alg:const} extracts subtrees from a constituent parse tree. These subtrees are \textit{constituents}, and break down sentences into meaningful spans of text. Constituents are one of

\begin{enumerate}
    \setlength\itemsep{-.4em}
    \item A relative clause
    \item The highest level S or SBAR node in its subtree with (NP, VP) children
    \item The highest level VP node above 2)
    \item The remaining nodes in the tree that were not extracted with 1), 2) or 3)
\end{enumerate}

\begin{algorithm*}[ht]
    \DontPrintSemicolon
    \caption{\textsc{ConstituentSegments}}
    \label{alg:const}
      
    \KwIn{sentence parse tree \var{PT}}
    $\var{const\_subtrees} := [\;]$  \tcp*{Store constituent subtrees here}
    $\var{PT}, \var{punct\_idxs} := \textsc{RemovePunct}(\var{PT})$\\
    \ForEach(\tcp*[f]{Find all constituent subtrees.}){subtree \var{ST} in \var{PT}}
    {
        \If{\var{\{NP, VP\}} in \var{ST.children}}
        {
            $\var{STAG} = \var{ST}$\\
            \tcc{Ascend as far as possible in tree, before root S tag}
            \While{\var{STAG.parent} in \var{\{SBAR, S, VP\}} \ttcmd{and} \var{STAG.parent} != \var{PT.root}}
            {
                $\var{STAG} := \var{STAG.parent}$\\
                \If{\var{STAG} in \var{\{S, SBAR\}} \ttcmd{and} not \var{STAG.children.intersection(\{VP, NP\})}}
                {\ttcmd{break}} 
                \tcc{If \var{STAG} is a VP, no need to break}
            $\var{const\_subtrees} := \var{const\_subtrees} + \var{[STAG]}$
            }
        }
        \ElseIf{\textsc{IsRelativeClause(\var{ST})}}
        {
            $\var{const\_subtrees} := \var{const\_subtrees} + \var{[STAG]}$
        }
    }
    \tcc{Create words list for each constituent subtree. Avoid duplicating words by removing subtrees that we add from the original parse tree.}
    \ForEach{subtree \var{ST} in \var{const\_subtrees} }
    {
        $\var{WORDS} := [\;]$ \tcp*{constituent word lists}
        
        \ForEach(\tcp*[f]{Break up clauses of conjunctions.}){\var{left} in \var{ST.left\_siblings}}
        {
            \If {$\var{left} = \var{CC}$} {
                $\var{WORDS} := \var{WORDS} + [\var{left.words}]$\\
                \textsc{RemoveSubtree}(\var{PT}, \var{left})
            }
        }
        $\var{WORDS} := \var{WORDS} + [\var{ST.words}]$\\
        \textsc{RemoveSubtree}(\var{PT}, \var{ST})
    }
    \If(\tcp*[f]{Add any remaining words to another segment}){\var{PT.words}}
    {
        $\var{WORDS} := \var{WORDS} + [\var{PT.words}]$
    }

    \var{WORDS} := \textsc{SplitNoncontiguous}(\var{WORDS})\\
    \var{WORDS} := \textsc{SortByIndex}(\var{WORDS})\\
    \var{WORDS} := \textsc{InsertPunctuation}(\var{WORDS}, \var{punct\_idxs})\\
    \var{WORDS} := \textsc{ConcatenateShortSegments}(\var{WORDS})\\
    \var{constituents} := \textsc{JoinWordLists}(\var{WORDS})\\
    \KwOut{constituents $c_1, ..., c_n$}
\end{algorithm*}

\begin{figure*}[t!]
\centering
\includegraphics[width=\textwidth]{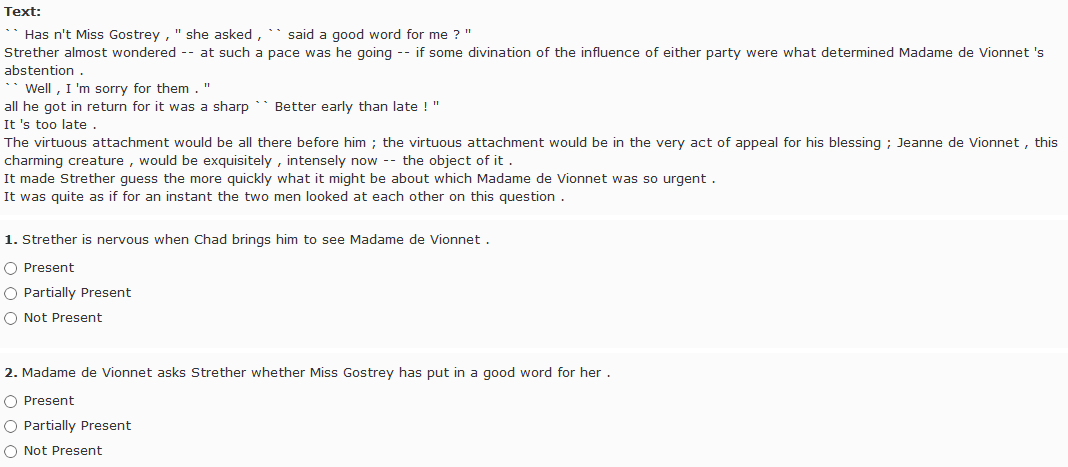}
\caption{An example HIT showing a segmented oracle summary, and two questions. Reading the summary, we see that we should answer "Present" for both questions. There can be up to 12 questions -- we omit here for brevity. Note that in our evaluation, we counted both "Present" and "Partially Present" as a match.}
\label{fig:hit}
\end{figure*}
\end{document}